\documentclass[11pt]{article}

\usepackage[final]{acl}

\usepackage{times}
\usepackage{latexsym}

\usepackage[T1]{fontenc}

\usepackage[utf8]{inputenc}

\usepackage{microtype}

\usepackage{inconsolata}

\usepackage{graphicx}

\usepackage{amssymb}
\usepackage{amsmath}
\usepackage{float}
\usepackage{listings}
\usepackage[hang]{footmisc}
\title{\textsc{NotAI.AI}: An Interactive Explainable System for\\AI-Generated Text Detection}

\author{
    \textbf{Oleksandr Marchenko Breneur\textsuperscript{*}} \quad \textbf{Adelaide Danilov\textsuperscript{*}} \\
    \textbf{Aria Nourbakhsh} \quad \textbf{Salima Lamsiyah} \\
    Department of Computer Science, \\
    Faculty of Science, Technology and Medicine, University of Luxembourg \\
    \small{\textbf{Correspondence:} \href{mailto:oleksandr.marchenko.002@student.uni.lu}{oleksandr.marchenko.002@student.uni.lu}} \\
    \small\textsuperscript{*}Equal contribution
}

\begin{document}
\maketitle

\begin{abstract}
We present \textsc{NotAI.AI}, an explainable AI-generated text detection system. Instead of returning only a binary label or confidence score, the system shows which signals influenced the prediction and lets users inspect an attribution-based sensitivity estimate obtained by subtracting selected local contributions. \textsc{NotAI.AI} combines sentence-level conditional probability curvature, a neural detector score, and interpretable stylometric and readability features in an XGBoost meta-classifier. It explains predictions with TreeSHAP feature contributions and can turn the resulting evidence into a concise natural-language explanation. We evaluate the system on a category-balanced subset of RAID containing human-written, clean AI-generated, and attacked AI-generated texts. The full model outperforms variants based on individual feature families, reaching 0.9685 F1 on the held-out within-subset test split. In an automatic evaluation, two model judges rate 94.5-98.6\% of generated explanations as faithful to the supplied detector evidence. The web interface\footnote{\url{https://notai-ai.vercel.app}}, source code\footnote{\url{https://github.com/Oleksandr-MB/EMNLP2026DEMO_NotAI.AI}}, and demonstration video\footnote{\url{https://youtu.be/l8Nk8kdBTHQ}} are publicly available.

\emph{The demo is deployed with an on-demand GPU server. After periods of inactivity, the first request may require a backend initialization.}

\end{abstract}

\section{Introduction}

Large language models (LLMs) have made fluent text generation widely available across education, journalism, research, and everyday writing \cite{naveed2023comprehensive,chang2024survey,chung-etal-2023-increasing}. This has clear benefits, but it also makes the origin of a text harder to judge. A student may revise an AI-generated draft, a journalist may receive generated submissions, or a reviewer may need to assess whether a text has been heavily machine-written. In such cases, the question is often not only whether a detector predicts ``AI'' or ``human'', but why it makes that prediction.

Most deployed AI-text detectors provide little to no help with the second question. They usually return a label, a score, or both. However, a score by itself is difficult to interpret, especially when detectors are known to be sensitive to domain shifts, decoding choices, paraphrasing, and adversarial transformations \cite{dugan-etal-2024-raid,wu2025survey}. A useful detection interface should therefore expose the evidence behind its decision, so that users can see whether the prediction is supported by several signals or mainly driven by one fragile cue.

We introduce \textsc{NotAI.AI}, a web-based demo system for explainable machine-generated text detection. The system combines three types of evidence: sentence-level conditional probability curvature, a neural detector score, and interpretable stylometric and readability features. These features are passed to an XGBoost meta-classifier \cite{chen2016xgboost}, and the resulting prediction is explained using TreeSHAP feature contributions \cite{lundberg2017unified}.

The interface is designed around this explanation. Users can paste text, inspect the predicted label and AI probability, compare the strongest human-supporting and AI-supporting features, and run an attribution-sensitivity analysis that subtracts selected local contributions from the model logit. The system can also generate a concise natural-language explanation from the feature values and their contributions.

Our contribution is primarily at the system and interaction-design level rather than a new detection or explanation algorithm: we integrate established detector signals and make their competing local contributions inspectable in one workflow.

We evaluate \textsc{NotAI.AI} on a category-balanced subset of RAID containing human-written, clean AI-generated, and attacked AI-generated texts \cite{dugan-etal-2024-raid}. The full model outperforms variants that use only stylometric, neural, or sentence-level curvature features, showing that these feature families provide complementary signals within this subset. The demo shows users these signals directly, rather than providing a single detection score.

\section{Related Work}

\paragraph{Machine-generated text detection.}
AI-generated text detection has been studied through a range of statistical, neural, and watermarking-based methods \cite{wu2025survey}. Early user-facing approaches such as GLTR expose token-level likelihood patterns to help users inspect whether a text follows model-like distributions \cite{gehrmann-etal-2019-gltr}. Other work uses likelihood-based signals more directly: DetectGPT identifies generated text through curvature in the probability landscape under perturbations \cite{mitchell2023detectgpt}, while Fast-DetectGPT estimates a related curvature signal more efficiently through conditional probability statistics \cite{bao2023fast}. Supervised neural detectors, including Transformer-based classifiers such as RoBERTa, provide another strong family of methods and are commonly evaluated across multi-domain and multi-generator benchmarks \cite{liu2019roberta,wang-etal-2024-m4}. Watermarking methods instead embed detectable patterns during generation \cite{liu2024survey,kirchenbauer2023watermark}, but they require cooperation from the generator and can be affected by paraphrasing or post-editing.

\paragraph{Robustness and interpretability.}
A key challenge for AI-text detectors is robustness. Detectors that perform well in one setting can degrade under changes in domain, generator, decoding strategy, or attack type \cite{dugan-etal-2024-raid}. This makes explanation important: when a detector gives a prediction, users should be able to inspect which signals drove it rather than relying only on a score. Prior systems such as GLTR and LLM-DetectAIve show the value of presenting detector evidence in a user-facing interface \cite{gehrmann-etal-2019-gltr,abassy2024llm}. Recent work has also studied feature-level patterns in AI-text detection \cite{kuznetsov-etal-2025-feature} and more complex authorship settings such as human-AI co-writing \cite{su-etal-2025-haco}.

\textsc{NotAI.AI} follows this line of work by combining several complementary detector signals and exposing their contributions to the user. Unlike systems that return only a final label, our system shows both AI-supporting and human-supporting evidence and lets users test how selected feature contributions affect the prediction.

\section{\textsc{NotAI.AI} System Overview}\label{sec:system-overview}
\begin{figure}[t!]
    \centering
    \includegraphics[width=1\linewidth]{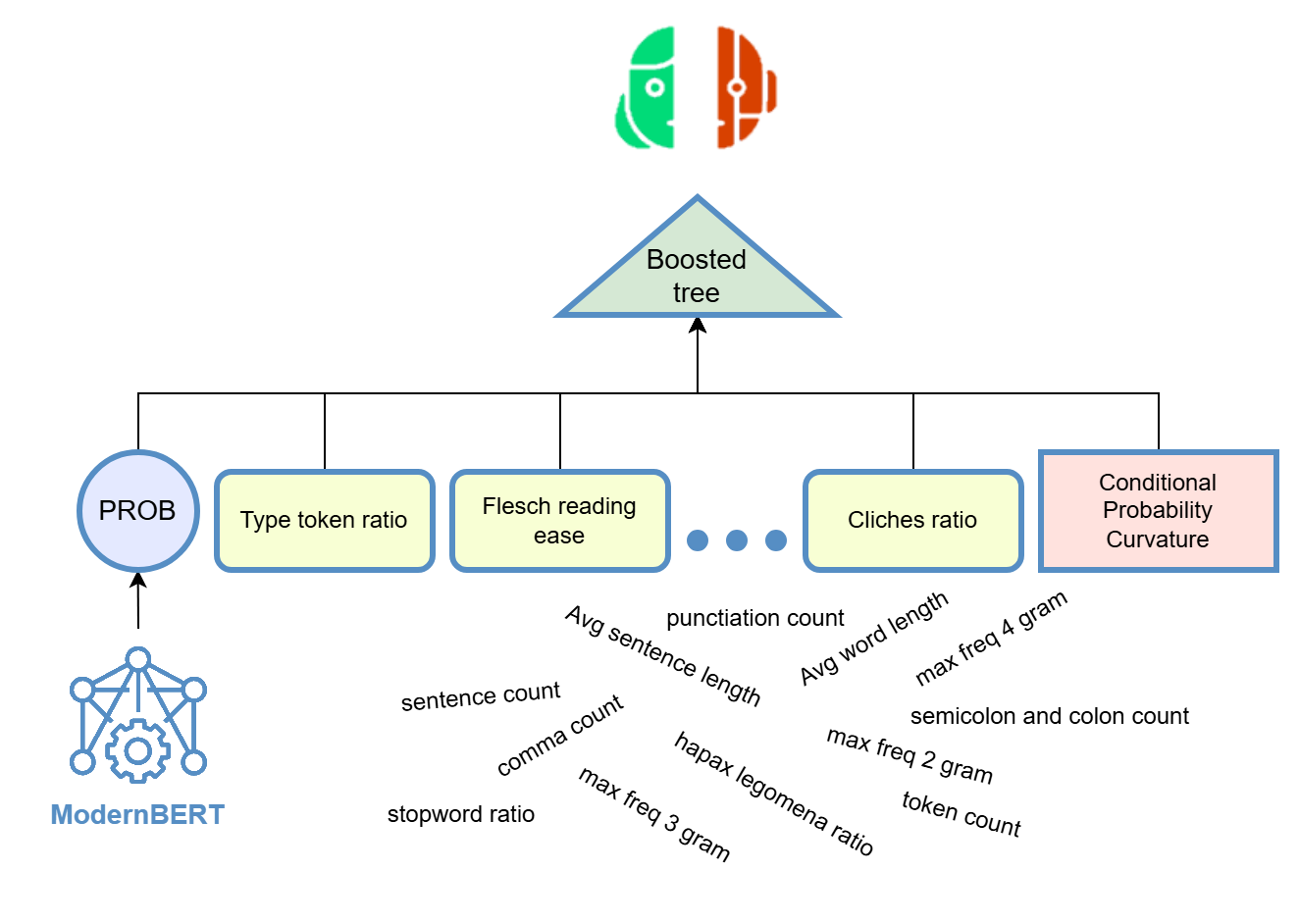}
    \caption{\textsc{NotAI.AI} classifier architecture.}
    \label{fig:architecture}
\end{figure}

\textsc{NotAI.AI} takes an input text and returns both a detection result and an explanation of the evidence behind it. The system has four main stages: feature extraction, classification, attribution, and presentation. First, the backend extracts neural, curvature-based, readability, and stylometric features from the input text. These features are passed to an XGBoost meta-classifier, which predicts whether the text is human-written or AI-generated. The prediction is then explained with feature-level contributions and shown in the web interface together with the strongest evidence for each class. \autoref{fig:architecture} shows the overall classifier pipeline.

\subsection{Feature Extraction}\label{sec:feature-extraction}

\paragraph{Neural detector score.}
The neural signal comes from a fine-tuned ModernBERT detector\footnote{\url{https://huggingface.co/GeorgeDrayson/modernbert-ai-detection}} \cite{warner2025smarter,drayson2025machine}. The checkpoint's model card reports that it was trained on the MAGE dataset; RAID is not listed as its training source. We use this third-party checkpoint without further fine-tuning. The web interface limits input to 512 words, while the neural detector tokenization truncates input to 512 tokens. We use the resulting softmax AI-class score as the feature \texttt{bert\_ai\_score}; we do not assume that this score is calibrated. It is one input to the meta-classifier rather than a standalone verdict, and its effect on a particular prediction is represented by its local contribution.

\paragraph{Sentence-level conditional probability curvature.}
The second signal is based on Conditional Probability Curvature (CPC), following Fast-DetectGPT \cite{bao2023fast}. CPC compares the likelihood of the observed text with the expected likelihood under a proxy language model. In practice, it captures whether the text lies in a smoother and more predictable region of the model distribution, which is often associated with generated text.

Instead of using a single document-level CPC score, \textsc{NotAI.AI} computes CPC over sentence spans. The final feature set includes six sentence-level CPC features: \texttt{cpc\_sent\_mean}, \texttt{cpc\_sent\_std}, \texttt{cpc\_sent\_min}, \texttt{cpc\_first\_sent}, \texttt{cpc\_middle\_sent}, and \texttt{cpc\_last\_sent}. These features let the classifier use both the overall curvature profile of the text and position-specific information from the beginning, middle, and end.

\paragraph{Readability and lexical features.}
We also extract features that describe the structure and lexical diversity of the text. These include \texttt{token\_count}, \texttt{sentence\_count}, \texttt{avg\_sentence\_length}, \texttt{avg\_word\_length}, \texttt{type\_token\_ratio}, \texttt{hapax\_legomena\_ratio}, and \texttt{flesch\_reading\_ease}. These features capture simple but useful properties such as text length, sentence structure, word choice, and readability.

\paragraph{Surface and stylometric features.}
Finally, the system extracts surface-level writing cues: \texttt{max\_freq\_2gram}, \texttt{max\_freq\_3gram}, \texttt{max\_freq\_4gram}, \texttt{punctuation\_count}, \texttt{comma\_count}, \texttt{semicolon\_and\_colon\_count}, \texttt{cliches}, and \texttt{stopword\_ratio}. These features measure repetition, punctuation use, formulaic phrasing, and function-word frequency. Several of them are normalized by token count so that texts of different lengths can be compared more fairly.

\begin{figure*}[t!]
    \centering
    \includegraphics[width=0.9\linewidth]{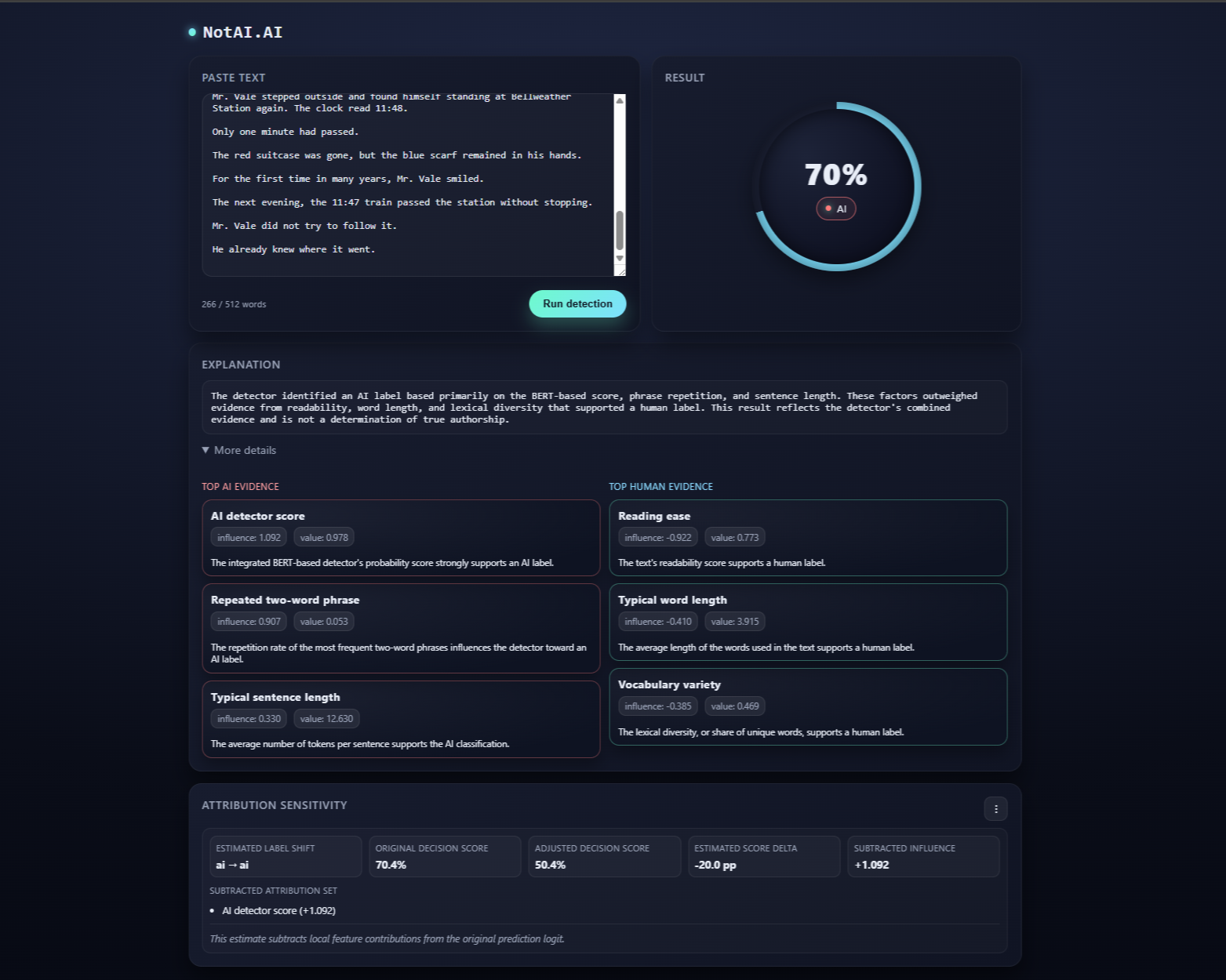}
    \caption{\textsc{NotAI.AI} user interface. Sample text generated using \texttt{ChatGPT} \cite{openai2023chatgpt}.}
    \label{fig:ui}
\end{figure*}

\subsection{Meta-Classification}

The extracted feature vector is passed to an Extreme Gradient Boosting classifier implemented with XGBoost \cite{chen2016xgboost}. The full model uses 22 features: 15 stylometric and readability features, 6 sentence-level CPC features, and the neural detector score. The classifier outputs an AI probability, and the final label is obtained by thresholding this probability.

A tree-based meta-classifier is suitable here because the feature families capture different types of evidence and need not combine linearly. For example, a high neural detector score may be more convincing when it agrees with curvature and stylometric evidence, while the same score may be less reliable when other features point in the opposite direction.

\subsection{Explainability}

For each prediction, \textsc{NotAI.AI} computes additive feature contributions using the TreeSHAP implementation for XGBoost \cite{lundberg2017unified}. The contributions are defined in logit space: positive values push the prediction toward the AI class, while negative values push it toward the human class. The interface shows the strongest AI-supporting and human-supporting features together with their raw feature values.

The system can also generate a short natural-language explanation using \texttt{google/gemma-4-31b-it} \cite{gemmateam2026gemma4technicalreport} through OpenRouter, prompted as shown in \autoref{sec:prompt}. The explainer receives the predicted label, AI probability, raw feature values, and the strongest positive and negative feature contributions. We also provide secondary wording guidance derived from RAID-based SHAP dependence plots. Because these value ranges are dataset-specific and may not transfer to arbitrary pasted text, the prompt treats the supplied local contribution as authoritative, forbids provenance claims, and describes the output only as evidence used by this detector.

\subsection{Demo Workflow}

Users interact with \textsc{NotAI.AI} through a web interface, shown in \autoref{fig:ui}. They paste text into the input panel and receive a predicted label, an AI probability, and a ranked list of features that influenced the prediction. The interface separates evidence supporting the AI class from evidence supporting the human class, which helps users see whether the decision is one-sided or mixed.

The demo also includes an attribution-sensitivity panel. Given a prediction logit $\ell$ and a set of selected local feature contributions $\varphi_i$, the interface reports an adjusted logit $\ell' = \ell - \sum_i \varphi_i,$ and converts it back into an AI probability. This view shows the additive effect assigned to those contributions for the current prediction. It does not rerun or retrain the model with the features absent, does not recompute tree interactions, and therefore must not be interpreted as a counterfactual model prediction. The interface displays this caveat next to the adjusted score.

\section{Evaluation}

\subsection{Dataset}
We evaluate \textsc{NotAI.AI} on a category-stratified subset of the RAID benchmark \cite{dugan-etal-2024-raid}. We selected RAID because it jointly provides the benchmark coverage and metadata required for our controlled evaluation, including multiple domains, generators, decoding settings, adversarial transformations, and source relationships. The detector features themselves are computed by our pipeline rather than supplied by RAID. The subset contains $40{,}000$ RAID training examples: $13{,}333$ clean human (99.7\% of the training partition's $13{,}371$), $13{,}333$ clean AI, and $13{,}334$ attacked AI. It is therefore approximately balanced across these three source categories, not across the binary labels: the two AI categories together make the binary task approximately 2:1 AI to human. RAID does not provide attacked-human counterparts; its adversarial transformations are defined for machine-generated texts.

The clean AI portion is balanced across the 11 generator labels, with approximately the same number of examples per generator. The attacked AI portion is balanced across generator and attack type, so that the evaluation does not depend on a single generator or transformation type. This setup supports standard human-vs-AI detection on clean text and measures detection of attacked AI within this controlled subset.

\subsection{Experimental Settings}

For each example, we extract the feature set described in  \autoref{sec:feature-extraction}. This includes sentence-level CPC features, the ModernBERT detector score, and stylometric, lexical, and readability features. The feature table is also released for inspection on Hugging Face\footnote{\url{https://huggingface.co/datasets/chemige/RAID_Balanced}}.

All model variants are trained and evaluated on the same dataset split, using the same precomputed feature values. We divide the data into approximately 85\% training, 5\% validation, and 10\% test examples. To avoid placing related texts (for example, text $t$ and its attacked version $t'$) in different splits, we group the examples by their original source ID. The split is also stratified by label. We tune the decision threshold on the validation split and report the main results on the held-out test split. The CPC features are computed with \texttt{EleutherAI/gpt-neo-1.3B} as the proxy language model \cite{gpt-neo,gao2020pile}. The proxy model is not assumed to be the generator of the evaluated text; it is used only to estimate local likelihood behavior. For all feature-family ablations, we train XGBoost classifiers using the same training procedure, except for the feature set available to the classifier.

The split holds out source groups but not entire generators, domains, or attack types. Consequently, the results measure in-distribution performance within this constructed RAID subset; they are not directly comparable to the official RAID evaluation protocol and do not establish robustness to unseen generators, domains, or attacks.

\subsection{Classification Evaluation}

\autoref{tab:raid_results} reports the main held-out test results. Because the contribution of \textsc{NotAI.AI} is the explainability workflow rather than a claim of state-of-the-art detection performance, the classification evaluation focuses on whether the integrated feature families provide complementary evidence. We compare the full ensemble against three ablations that use only one feature family: stylometric and readability features, the ModernBERT detector score, or sentence-level CPC features. All variants use the same train/validation/test split and the same threshold-selection procedure described above. Precision, recall, and F1 are reported for the AI class.

\begin{table}[t!]
\centering

\resizebox{\columnwidth}{!}{%
\begin{tabular}{lccccc}
\hline
    \textbf{Model} & \textbf{Acc.} & \textbf{Prec.} & \textbf{Rec.} & \textbf{ROC AUC} & \textbf{F1} \\
    \hline
    All-AI majority & 0.6650 & 0.6650 & 1.0000 & 0.5000 & 0.7988 \\
    Stylometric only & 0.8895 & 0.9117 & 0.9233 & 0.9511 & 0.9174 \\
    ModernBERT only & 0.8700 & 0.8749 & 0.9387 & 0.9122 & 0.9057 \\
    Sentence CPC only & 0.8562 & 0.9306 & 0.8470 & 0.9246 & 0.8868 \\
    Full ensemble & \textbf{0.9583} & \textbf{0.9731} & \textbf{0.9639}  & \textbf{0.9924} & \textbf{0.9685} \\
    \hline
    \end{tabular}
}
\caption{Held-out test performance on the category-balanced RAID subset. Precision, recall, and F1 are computed for the AI class.}
\label{tab:raid_results}
\end{table}

The full ensemble performs best across all reported metrics, reaching 0.9685 F1 and 0.9924 ROC AUC. This is a large improvement over both the 0.7988 F1 all-AI majority baseline and the strongest single-family variant, the stylometric-only model, which reaches 0.9174 F1. On the 1,340 human test examples, the full ensemble correctly classifies 1,269 and produces 71 false positives, corresponding to 0.9470 human-class recall (specificity), 0.9383 human-class F1, and a 5.3\% false-positive rate. The results suggest that the three feature families capture complementary signals: stylometric features are strong on their own, ModernBERT provides high recall, and sentence-level CPC provides high precision but lower recall.

The clean and attacked subsets also show different behavior across feature families. ModernBERT performs especially well on attacked AI texts, while CPC is more reliable on clean text than on attacked text. The full ensemble reaches 0.9603 accuracy on clean test examples and 0.9540 accuracy on attacked AI examples. Because all attacked examples are AI-labeled, the latter number may partly reflect transformation artifacts and must not be interpreted as general attack robustness. We include the detailed F1 heatmap per topic-generator pair in \autoref{fig:f1} (see \autoref{sec:f1}).

\subsection{Explanation Evaluation}

We evaluate the faithfulness of the generated explanations on a category-stratified sample of 500 examples from the complete RAID-derived feature table (167 human, 167 clean AI, and 166 attacked AI). For each example, \texttt{google/gemma-4-31b-it} receives the detector prediction, probability, raw feature values, and strongest local TreeSHAP contributions. We then evaluate the resulting explanation against this supplied evidence using two separate model judges, \texttt{google/gemma-4-26b-a4b-it} \cite{gemmateam2026gemma4technicalreport} and \texttt{qwen/qwen3.6-35b-a3b} \cite{qwen36-35b-a3b}. The judges assess whether feature directions and values are faithful, whether the summary is consistent with the prediction, whether it introduces no unsupported claims, and whether the explanation avoids claiming true authorship. We define an explanation as overall faithful only when all five criteria pass. Thus, this evaluation measures fidelity to the detector evidence rather than whether the detector's prediction is correct.

All 500 explanations were valid JSON. The deterministic checks found no unexpected features and no polarity errors; mean coverage of the supplied evidence was 98.4\%, and 98.4\% of explanations contained a non-empty summary. TreeSHAP contributions also closely reconstructed the detector logit, with a mean absolute error of $3.26\times10^{-6}$ and a maximum error of $1.99\times10^{-5}$.

\begin{table}[t!]
\centering
\resizebox{\columnwidth}{!}{%
\begin{tabular}{lcccc}
    \hline
    \textbf{Criterion} & \textbf{Gemma (\%)} & \textbf{Qwen (\%)} & \textbf{Agree (\%)} & $\boldsymbol{\kappa}$ \\
    \hline
    Polarity correct & 98.8 & 95.4 & 94.9 & 0.130 \\
    Values faithful & 98.6 & 94.5 & 93.8 & 0.103 \\
    Summary consistent & 98.8 & 98.9 & 98.5 & 0.356 \\
    No unsupported claims & 98.8 & 94.5 & 94.1 & 0.110 \\
    No authorship claim & 98.8 & 99.1 & 98.7 & 0.394 \\
    Overall faithful & 98.6 & 94.5 & 93.8 & 0.103 \\
    \hline
\end{tabular}
}
\caption{Automatic explanation-faithfulness results. The Gemma judge produced 500 valid judgments, and Qwen produced 454; agreement and Cohen's $\kappa$ are computed over the 454 paired judgments. Overall faithfulness requires all five component criteria to pass.}
\label{tab:explanation_eval}
\end{table}

As shown in \autoref{tab:explanation_eval}, both judges rate more than 94\% of their valid judgments as faithful overall. Their overall decisions agree on 93.8\% of paired examples ($\kappa=0.103$). Agreement is strongest for avoiding authorship claims (98.7\%, $\kappa=0.394$), while the remaining criteria have raw agreement between 93.8\% and 98.5\% and lower chance-corrected agreement. Qwen is somewhat more conservative in its error counts, reporting 0.139 contradictions and 0.042 unsupported claims per valid judgment on average, compared with 0.002 and 0.000 for Gemma. Qwen produced 454 valid JSON judgments; its other 46 responses could not be parsed as JSON and are excluded from its rates and the agreement analysis. Overall, the results indicate that the explainer usually verbalizes the supplied local evidence correctly, although incomplete judge coverage and judge disagreements leave room for stricter output validation.

\subsection{Feature Influence Analysis}

\begin{figure*}[t!]
    \centering
    \includegraphics[width=0.9\linewidth]{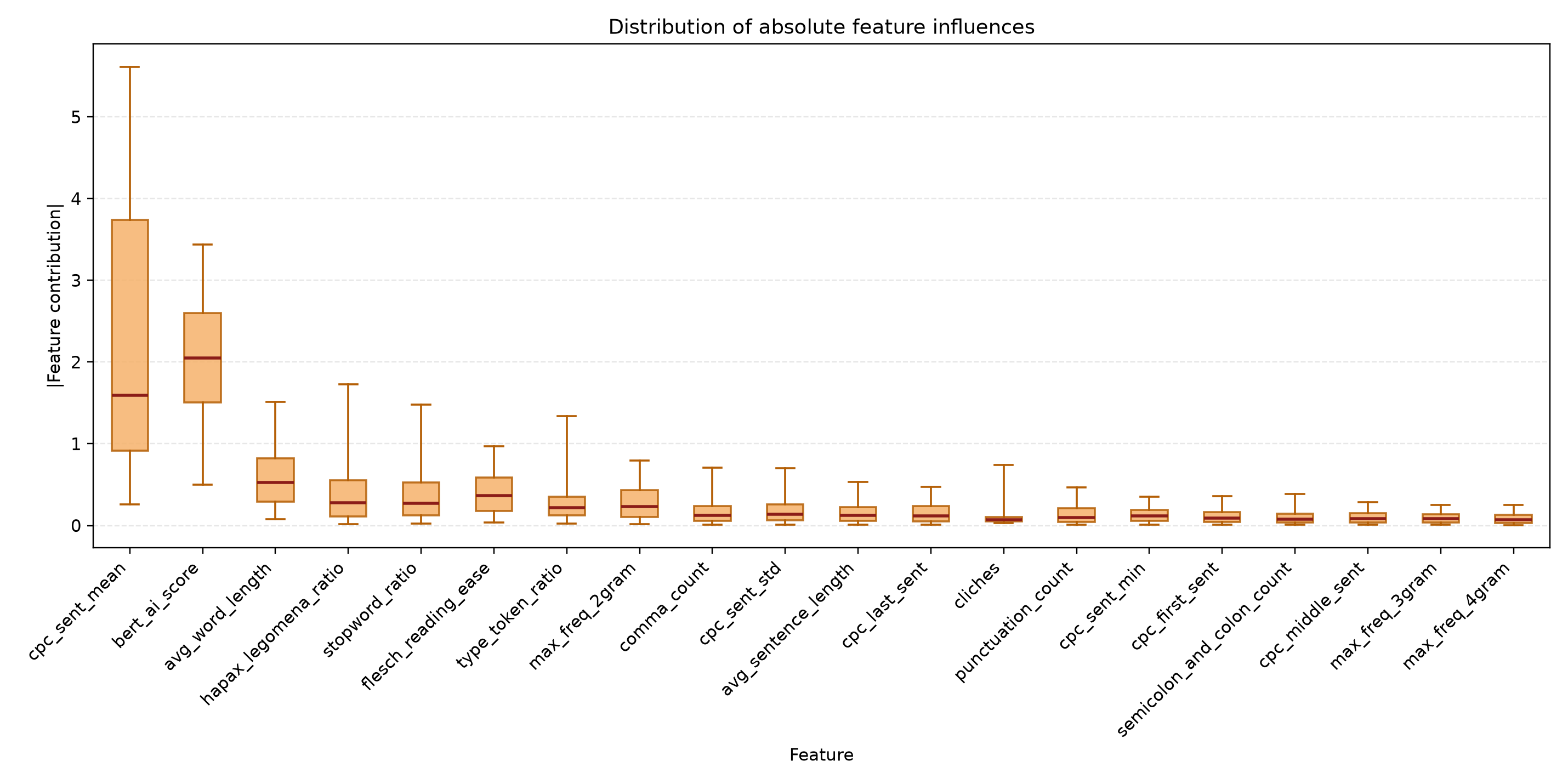}
    \caption{Distribution of absolute feature contributions for the full ensemble. Features are ordered by their average absolute contribution.}
    \label{fig:feature_importance}
\end{figure*}

\begin{figure}[t!]
    \centering
    \includegraphics[width=\linewidth]{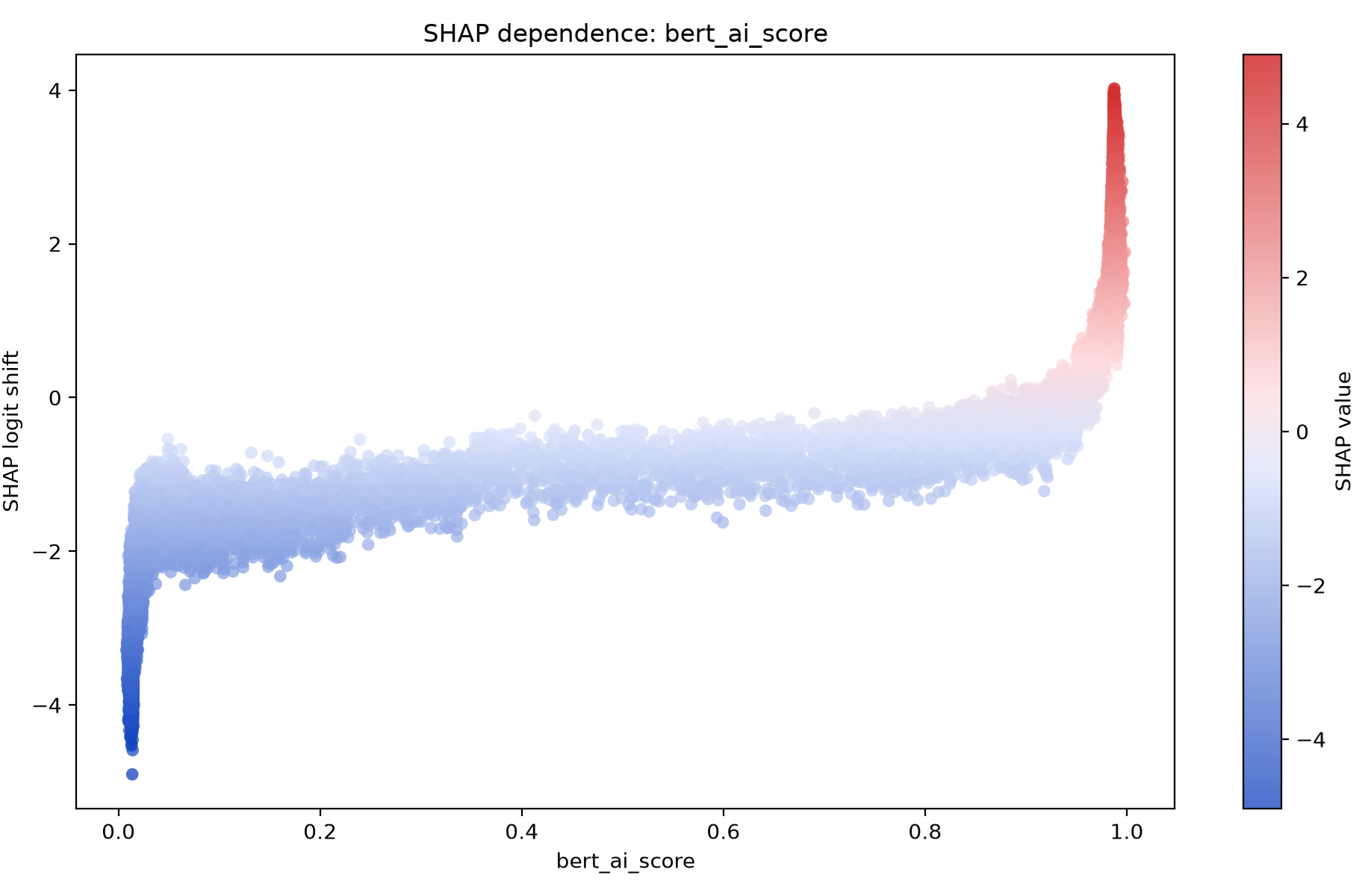}
    \caption{ModernBERT score dependence in the full ensemble. The axes show the raw neural score and its TreeSHAP contribution to the AI logit.}
    \label{fig:bert_shap}
\end{figure}

We collect absolute TreeSHAP contributions across the complete feature table in \autoref{fig:feature_importance}. \texttt{cpc\_sent\_mean} has the largest average absolute contribution and a wide spread, showing that curvature is central but highly instance-dependent.

\texttt{bert\_ai\_score} is the second strongest feature. Its raw distribution is polarized: median scores are 0.0281 for clean human, 0.9873 for clean AI, and 0.9868 for attacked AI. Nevertheless, 27.6\% of human examples score at least 0.5, while 5.8\% of clean AI and 7.5\% of attacked AI examples score below 0.5. \autoref{fig:bert_shap} shows that the ensemble uses the score nonlinearly: low and intermediate scores generally reduce the AI logit, while scores near the upper end can increase it sharply. The overlap confirms that this is an ensemble feature, not a standalone verdict.

The strongest interpretable style features are \texttt{avg\_word\_length}, \texttt{hapax\_legomena\_ratio}, \texttt{stopword\_ratio}, and \texttt{flesch\_reading\_ease}. However, \texttt{type\_token\_ratio}, \texttt{max\_freq\_2gram}, \texttt{comma\_count}, and \texttt{cpc\_sent\_std} have moderate influence, while longer $n$-gram repetition features contribute less. Hence, the exposed evidence combines curvature-based, neural, and human-readable stylistic signals.

\section{Conclusion}
We presented \textsc{NotAI.AI}, a system for explainable AI-generated text detection. The system combines sentence-level CPC features, a neural detector score, and stylometric and readability features in an XGBoost meta-classifier. Instead of showing only a label or confidence score, the interface exposes feature-level evidence for both the AI and human classes, provides an attribution-sensitivity view with an explicit non-counterfactual caveat, and can generate a short natural-language explanation grounded in the model's feature contributions.

Our experiments on a category-balanced RAID subset show that the full ensemble outperforms variants using only one feature family. This suggests that neural, curvature-based, and stylometric signals capture complementary aspects of AI-generated text within this controlled setting. Two automatic judges rate 94.5-98.6\% of generated explanations as faithful to the supplied detector evidence. More importantly for the demo setting, \textsc{NotAI.AI} makes these signals visible to the user, allowing the prediction to be inspected rather than treated as a single opaque detector score.

\section*{Limitations and Future Work}

\textsc{NotAI.AI} currently uses binary human-vs.-AI classification, although real authorship is often more nuanced. Future work will extend the system to mixed-provenance settings, such as AI-generated text that has been subsequently humanized or human-written text that has been lightly revised by AI.

Our evaluation is limited to English texts from a category-balanced subset of the RAID training partition. Because the split does not hold out entire generators, domains, or attack types, the results do not establish out-of-distribution robustness or generalization to newer models. Moreover, RAID contains no attacked-human counterparts, so performance on attacked AI text may partly reflect transformation artifacts. Broader evaluation should include unseen generators, domains, languages, attacks, and real-world editing workflows.

Classification comparisons are limited to a majority baseline and internal feature-family ablations, as our goal is to test the complementarity of the evidence exposed by the interface rather than establish state-of-the-art detection performance. Accordingly, the reported results should be interpreted as evidence of internal complementarity rather than benchmark superiority.

The explanation interface describes the model's decision process rather than establishing true authorship. TreeSHAP reports feature contributions to the XGBoost prediction, while the sensitivity panel subtracts selected additive attributions from the original logit and should not be interpreted as a counterfactual prediction with those features removed.

Finally, the natural-language explanations rely on local contributions and RAID-derived SHAP ranges that may not transfer out of distribution. Although our automatic evaluation measures faithfulness to the supplied detector evidence, model-based judges do not replace human annotation or demonstrate that the interface improves user judgments. Future work will explore stricter explanation validation and user studies.

\section*{Ethics Statement}

\textsc{NotAI.AI} is a transparency-oriented demo for inspecting evidence used in AI-generated text detection. Its outputs are model-based signals and should not replace human judgment.

Key risks include false positives, privacy of submitted text, and uneven performance across writing styles or user groups, particularly for students and non-native writers. To discourage over-reliance on a single score, the interface presents both AI-supporting and human-supporting evidence.

Submitted text should be treated as sensitive. The demo does not require identifying information. If the optional natural-language explanation is used, the text and detector evidence are sent through OpenRouter to the selected model provider. This data flow is disclosed to users, who should consult the selected provider's applicable retention policy.


\bibliography{custom}

\clearpage
\appendix

\onecolumn

\section*{Appendix}

\section{Explainer Prompt}\label{sec:prompt}

\begin{lstlisting}
Explain the AI detector's output from the supplied feature values and SHAP importance scores. Positive scores support the AI label; negative scores support the human label.

You will receive JSON with these keys:
- raw_text: text snippet or summary
- label: ai or human
- probability_ai: the float number with probability
- features_positive: the dictionary containing top features influencing the "AI" decision. Name of features are keys and dictionary with importance_score and raw_value are values for each feature key
- features_negative: the dictionary containing top features influencing the "human" decision. Name of features are keys and dictionary with importance_score and raw_value are values for each feature key

Feature glossary (use these exact feature keys in returned JSON, but explain them in plain language). The ranges below are descriptive cut-points derived from the system's RAID-based SHAP dependence plots; they are not universal thresholds and may not transfer to arbitrary pasted text. Use them only to describe the detector's behavior. The supplied importance_score is the authority for the direction and strength of this specific prediction, even when a raw value appears inconsistent with a range below.
- token_count: total token count, if this feature is present. Very short texts below about 171 tokens and medium-long texts around 342-484 tokens support AI. Mid-length texts around 171-314 tokens support human, and extremely long texts above about 484 tokens strongly support human.
- sentence_count: number of sentences, if this feature is present. Fewer than about 11 sentences supports human. Around 11 or more sentences supports AI, with strongest AI evidence around 14-25 sentences.
- cpc_sent_mean: average CPC across sentences. Values below about 0.31 support human. Values above about 0.31 support AI, above 0.45 is strong AI evidence, and above about 0.79 is very strong AI evidence.
- cpc_sent_std: variation in CPC across sentences. Values below about 0.82 usually support human. Values above about 0.91 support AI, above 1.07 is stronger AI evidence, and above about 1.36 is very strong AI evidence.
- cpc_sent_min: lowest sentence CPC. This is non-monotonic: extremely low values below about -3.4 support AI, values from about -3.4 to -0.8 usually support human, and values above about -0.8 support AI, especially above about -0.25.
- cpc_first_sent: CPC of the first sentence. This feature is mixed and should follow the importance_score closely: values below about -1.62 support human, around -1.62 to -1.07 can support AI, around -0.13 to 0.33 supports human, and values above about 0.99 increasingly support AI.
- cpc_middle_sent: CPC of the middle sentence. This feature is mixed and non-monotonic: values below about -0.91 are human-leaning, around -0.91 to 0.45 supports AI, around 0.45-1.57 supports human, and very high values above about 2.33 support AI.
- cpc_last_sent: CPC of the last sentence. Values below about 0.31 usually support AI, especially below about -1.0. Values above about 0.31 usually support human, strongest around 0.85-1.64.
- bert_ai_score: probability (0-1) from the integrated BERT-based detector. Scores below about 0.91 support human. Scores above about 0.91 support AI, above 0.98 is strong AI evidence, and above about 0.986 is very strong AI evidence.
- avg_sentence_length: tokens per sentence. Values below about 20 support AI. Values around 20-27 support human, 29-36 supports AI again, and values above about 36 are weak or mixed.
- avg_word_length: mean word length. Very short words below about 3.87 support AI, and long words above about 4.89 support AI, especially above 5.37. Middle values around 3.87-4.89 usually support human.
- type_token_ratio: lexical diversity, the share of unique words. Values below about 0.56 usually support human. Values above about 0.56 support AI, above 0.61 is stronger, and above 0.76 is very strong AI evidence.
- hapax_legomena_ratio: share of words occurring once. Values below about 0.37 usually support human. Values above about 0.37 support AI, above 0.41 is stronger, and above 0.63 is very strong AI evidence.
- max_freq_2gram: highest normalized 2-word phrase repetition. Values below about 0.019 usually support human. Values above about 0.019 support AI, and above about 0.025 is stronger AI evidence.
- max_freq_3gram: highest normalized 3-word phrase repetition. This feature is weaker and mildly non-monotonic: values below about 0.010 support human, values around 0.010-0.022 support AI, and values above about 0.022 are weak or mixed.
- max_freq_4gram: highest normalized 4-word phrase repetition. This feature is weaker and non-monotonic: values below about 0.0039 support human, values around 0.0039-0.0125 support AI, and high values above about 0.0156 support human.
- punctuation_count: fraction of tokens that are punctuation. Values below about 0.10 usually support human or are near-neutral. Values above about 0.10 support AI, with stronger AI evidence above about 0.128.
- comma_count: fraction of tokens that are commas. Low values below about 0.026 support human. Middle values around 0.026-0.085 are mixed, so follow the importance_score. Very high values above about 0.085 strongly support AI.
- semicolon_and_colon_count: fraction of ";" or ":" tokens. Zero or tiny values below about 0.0036 usually support human. Values around 0.0036-0.012 support AI, with strongest AI evidence around 0.005-0.012; values above about 0.012 are weaker or mixed.
- cliches: fraction of tokens in the cliche list. Zero or near-zero values below about 0.0028 usually support human. Values above about 0.0028 support AI, and above about 0.0054 is strong AI evidence.
- stopword_ratio: fraction of stopwords. Very low or zero values below about 0.054 strongly support AI. Values around 0.054-0.39 usually support human, 0.39-0.44 is mixed, and values above about 0.44 support AI, strongest around 0.49-0.56.
- flesch_reading_ease: readability scaled roughly to 0-1, where higher is easier to read. Values below about 0.61 usually support AI, especially around 0.22-0.50. Values around 0.65-0.83 support human, while values above about 0.83 are weaker or mixed.

Your task:
1) Explain, in 1-2 sentences each, how these feature values support the final label.
2) Keep it concise, non-technical, and grounded in the supplied values; do not invent data.
3) Use the exact supplied feature keys as JSON keys, but explain them with plain-language descriptions from the feature glossary.
4) Describe evidence as influence on this detector, never as proof of AI or human authorship. Do not claim that a feature value is inherently AI-like or human-like outside this prediction.
5) Do not introduce confidence claims, provenance claims, thresholds, feature values, or causal explanations that are absent from the input. If a glossary range conflicts with importance_score, follow importance_score and avoid citing the range.
6) In the summary, state that the label reflects the detector's combined evidence and is not a determination of true authorship.

Return JSON:
{
  "top_ai_evidence": {"feature1" : "explanation1", "feature2" : "explanation2", ... "featureN" : "explanationN"},
  "top_human_evidence": {"feature1" : "explanation1", "feature2" : "explanation2", ... "featureN" : "explanationN"},
  "summary": "2-3 sentence plain-language rationale"
}
\end{lstlisting}
\newpage
\section{F1 Score Heatmap per Topic-Generator Pair}\label{sec:f1}
\begin{figure}[H]
    \centering
    \includegraphics[width=0.8\linewidth]{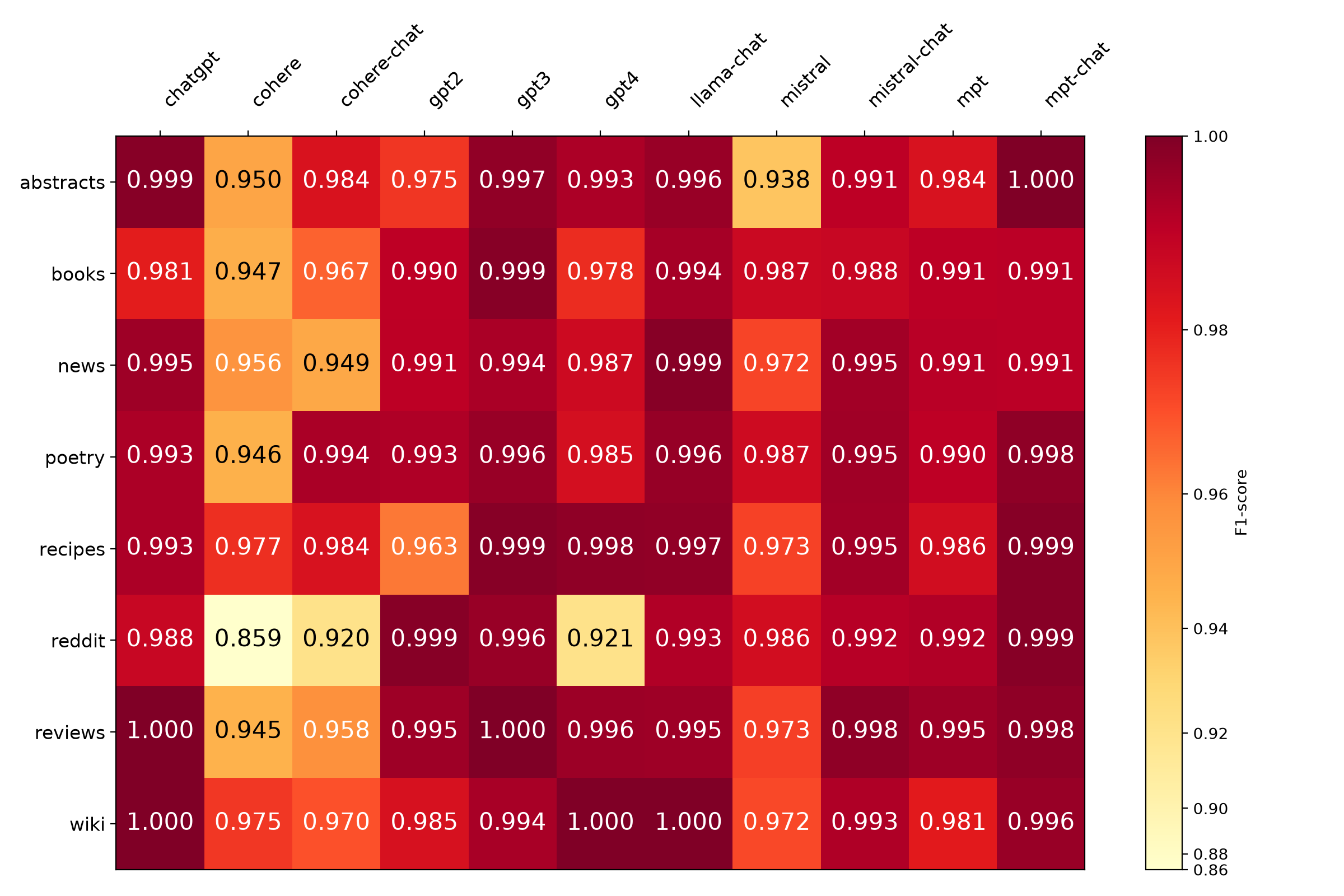}
    \caption{F1-Score heatmap per topic-generator model pair for the full ensemble classifier.}
    \label{fig:f1}
\end{figure}

\clearpage
\twocolumn
\end{document}